\documentclass[sigconf]{acmart}

\AtBeginDocument{%
  }

\copyrightyear{2025}
\acmYear{2025}
\setcopyright{cc}
\setcctype{by}
\acmConference[IVA Adjunct '25]{ACM International Conference on Intelligent Virtual Agents}{September 16--19, 2025}{Berlin, Germany}
\acmBooktitle{ACM International Conference on Intelligent Virtual Agents (IVA Adjunct '25), September 16--19, 2025, Berlin, Germany}\acmDOI{10.1145/3742886.3756709}
\acmISBN{979-8-4007-1996-7/2025/09}

\acmSubmissionID{37}

\usepackage{CJKutf8}

\usepackage{cleveref}
\usepackage{hyperref}
\usepackage{booktabs}
\usepackage{tabularx}
\usepackage{siunitx}
\usepackage{makecell}

\usepackage{caption}

\usepackage{color, colortbl}
\definecolor{darkgray}{gray}{0.85}
\definecolor{rowA}{rgb}{1,1,1}
\definecolor{rowB}{rgb}{0.95,0.95,0.95}
\definecolor{headingColor}{rgb}{0.6, 0.95, 0.7}
\definecolor{groupColor}{rgb}{0.90, 1.00, 0.90}
\definecolor{subtotalColor}{rgb}{0.80, 0.90, 1.00}
\definecolor{totalColor}{rgb}{0.50, 0.75, 0.95}
\begin{document}

\title{The TUB Sign Language Corpus Collection}

\author{Eleftherios Avramidis}
\orcid{0000-0002-5671-573X}
\affiliation{%
  \institution{German Research Center for AI (DFKI)}
  \department{Speech and Language Technology}
  \city{Berlin}
  \country{Germany}
}
\email{eleftherios.avramidis@dfki.de}

\author{Vera Czehmann}
\orcid{0009-0004-8007-6503}
\affiliation{%
  \institution{German Research Center for AI (DFKI)}
  \department{Speech and Language Technology}
  \city{Berlin}
  \country{Germany}
}
\affiliation{
  \institution{Technische Universität Berlin}
  \city{Berlin}
  \country{Germany}
}
\email{vera.czehmann@dfki.de}

\author{Fabian Deckert}
\orcid{0009-0003-5281-8226}
\affiliation{%
  \institution{Technische Universität Berlin}
  \city{Berlin}
  \country{Germany}
}
\email{fabian.deckert@campus.tu-berlin.de}

\author{Lorenz Hufe}
\orcid{0009-0000-4368-4877}
\affiliation{%
  \institution{Fraunhofer HHI}
  \city{Berlin}
  \country{Germany}
}
\affiliation{ 
  \institution{Bliss e.V.}
  \city{Berlin}
  \country{Germany}
  }
\email{lorenz.hufe@hhi.fraunhofer.de}

\author{Aljoscha Lipski}
\orcid{0009-0007-8672-6556}
\affiliation{%
  \institution{Technische Universität Berlin}
  \city{Berlin}
  \country{Germany}
}
\email{aljoscha.lipski@googlemail.com}

\author{Yuni Amaloa Quintero Villalobos}
\orcid{0000-0002-8569-1108}
\affiliation{%
  \institution{Technische Universität Berlin}
  \city{Berlin}
  \country{Germany}
}
\email{quintero.villalobos@campus.tu-berlin.de}

\author{Tae Kwon Rhee}
\orcid{0009-0001-0711-660X}
\affiliation{%
  \institution{Technische Universität Berlin}
  \city{Berlin}
  \country{Germany}
}
\email{tae.k.rhee@campus.tu-berlin.de}

\author{Mengqian Shi}
\orcid{0009-0005-8244-1735}
\affiliation{%
  \institution{Technische Universität Berlin}
  \city{Berlin}
  \country{Germany}
}
\email{mengqian.shi@campus.tu-berlin.de}

\author{Lennart Stölting}
\orcid{0009-0009-1104-0693}
\affiliation{%
  \institution{Technische Universität Berlin}
  \city{Berlin}
  \country{Germany}
}
\email{l.stoelting@campus.tu-berlin.de}

\author{Fabrizio Nunnari}
\orcid{0000-0002-1596-4043}
\affiliation{%
  \institution{German Research Center for AI (DFKI)}
  \department{Cognitive Assistants}
  \city{Saarbrücken}
  \country{Germany}
}
\email{fabrizio.nunnari@dfki.de}

\author{Sebastian Möller}
\orcid{0000-0003-3057-0760}
\affiliation{
  \institution{Technische Universität Berlin}
  \city{Berlin}
  \country{Germany}
}
\affiliation{
  \institution{German Research Center for AI (DFKI)}
  \department{Speech and Language Technology}
  \city{Berlin}
  \country{Germany}
}
\email{sebastian.moeller@tu-berlin.de}

\renewcommand{\shortauthors}{Avramidis, Czehmann, Deckert, Hufe, Lipski, Quintero Villalobos, Shi, Rhee, Stölting, Nunnari and Möller}

\begin{abstract}
We present a collection of parallel corpora of 12 sign languages in video format, together with subtitles in the dominant spoken languages of the corresponding countries. 
The entire collection includes more than 1,300 hours in 4,381 video files, accompanied by 1,3~M subtitles containing 14~M tokens.
Most notably, it includes the first consistent parallel corpora for 8 Latin American sign languages, whereas the size of the German Sign Language corpora is ten times the size of the previously available corpora.
The collection was created by collecting and processing videos of multiple sign languages from various online sources, mainly broadcast material of news shows, governmental bodies and educational channels.
The preparation involved several stages, including data collection, informing the content creators and seeking usage approvals, scraping, and cropping. 
The paper provides statistics on the collection and an overview of the methods used to collect the data.
\end{abstract}




\maketitle

\section{Introduction}

Language technology has made enormous progress over the last decade, resulting in several popular and helpful products and services.
However, it should be noted that while this progress has been observed for spoken languages, progress with regard to technology for sign languages has been limited.
As sign languages are the primary means of communication for deaf and many hard-of-hearing individuals, this lack of progress means that approximately half a billion people worldwide have fewer opportunities to experience the benefits of language technology, and therefore are disadvantaged with regard to communication and access to information and knowledge.
It should also be noted that for most sign language users, written language is a second language. This can lead to difficulties accessing most online content, which is predominantly textual, and limits benefits from text-based technologies.
We therefore believe that it is the responsibility of the language technology research community to encourage and empower research into relevant developments for sign languages, with the aim of increasing accessibility and lowering communication barriers around the world. 

A crucial element in the progress of language technology for spoken languages has been the focus on collecting and curating language corpora. This has facilitated research in fields such as \textit{computational linguistics} and \textit{natural language processing}. More recently, the conception and use of data-intensive empirical methods and machine learning has unlocked great potential in terms of functionality and applications, but this has depended on the availability of large amounts of corpora.
These corpora are predominantly textual and usually originate from publicly available web data that have been pre-processed and curated for specific tasks.
A relevant subcategory of these corpora are the so-called \textit{parallel corpora}, i.e. collections of content expressed in two or more languages in a way that can be aligned, allowing training of machine learning models for multilingual tasks, such as machine translation, multilingual summarization etc..

Unlike spoken languages, almost all sign languages suffer from a lack of corpora. 
We believe this is one of the reasons why the research and development of sign language technologies has been limited.
Although significant efforts have been made in recent years, with several datasets being released, these still only cover a few sign languages and are nowhere near what is required to train large-scale models similar to those available for spoken languages.

In an effort to improve this situation, we present the TUB sign language corpus collection: a set of parallel corpora sourced from the web containing videos of various sign languages alongside textual transcriptions of the dominant spoken languages of the countries where the respective SLs are used.
The important contributions of this collection are the following:
\begin{itemize}
\item the entire collection comprises 1,391 hours of continuous sign language videos, totalling 440 GB across 4,381 files,
\item the videos are accompanied by 1,335,320  timed closed captions (subtitles) in the respective spoken language (manually or automatically generated), in overall including 14 M tokens,
\item the corpora are manually collected from internet sources by 8 native speakers of the respective spoken languages,
\item the licences of the content have been verified, the content creators have been informed, and additional permissions have been sought and obtained by the majority of them,
\item the vast majority of the content consists of the interpretation of spoken televised or streamed content,
\item the content mostly falls in the domains of news reporting, government announcements and educational material
\item it includes the first parallel corpora of continuous sign language for 8 sign languages of Latin America. 
\end{itemize}
The collection is available open source in form of a catalogue, which is licensed under the MIT licence\footnote{\url{https://github.com/DFKI-SignLanguage/TUB-Sign-Language-Corpus-Collection}}.

The rest of the paper is structured as follows: 
In \Cref{sec:related_work} we provide an overview of the state-of-the-art sign language corpora and outline how our work compares to them. 
The construction of the collection is detailed in \Cref{sec:construction}. 
The content of the collection is described in \Cref{sec:corpora_description} whereas our view on the limitations is given in \Cref{sec:limitations} and a conclusion is provided in \Cref{sec:conclusion}. 

\begin{table*}[]
\centering
\begin{tabular}{lccrl}
\toprule
\textbf{Corpus Name} & \textbf{Sign Language} & \textbf{Spoken Language} & \textbf{Duration (hrs)} & \textbf{Annotation Granularity} \\
\midrule
Youtube-SL-25        & Various (25)          & Various                  & 3,207                  & Sentences                  \\
\arrayrulecolor{darkgray}\midrule
Youtube-ASL          & ASL                   & English                  & 1,000                  & Sentences                 \\
How2Sign             & ASL                   & English                  & 80                      & Sentences, glosses   \\
FLEURS-ASL           & ASL                   & English                  & 15                      & Sentences                 \\
\arrayrulecolor{darkgray}\midrule
DGS Corpus           & DGS                   & German                   & 50                     & Sentences, glosses  \\
RWTH-PHOENIX         & DGS                   & German                   & 13                      & Sentences, glosses             \\
DGS-Fabeln1          & DGS                   & German                   & 3                       & Sentences              \\
\arrayrulecolor{darkgray}\midrule
Matignon-LSF         & LSF                   & French                   & 39                      & Sentences   \\
MEDIAPI-SKEL         & LSF                   & French                   & 27                      & Sentences   \\
Dicta-Sign-LSF-v2    & LSF                   & French                   & 11                      & Sentences, glosses             \\
Rosetta-LSF          & LSF                   & French                   & 2                       & Sentences, glosses             \\
\arrayrulecolor{darkgray}\midrule
CORLSE               & LSE                   & Spanish                  & 380                    & Sentences  \\
\arrayrulecolor{darkgray}\midrule
KSL dataset          & KSL                   & Korean                   & 20                      & Sentences, glosses  \\ 
\arrayrulecolor{black}
\bottomrule
\end{tabular}
\caption{An overview of sign language datasets, sorted by language.}
\label{table:related_work}
\end{table*}

\section{Related Work}
\label{sec:related_work}

In this section we provide an overview of the state-of-the-art corpora for the sign languages included in our collection. An overview is given in \Cref{table:related_work}.

YouTube-SL-25 \cite{tanzerzhang-2024-youtubesl25} is the biggest and most multilingual parallel corpus between sign language videos with spoken language text. 
It contains solid representation of 25 languages and amounts to 3,207 hours of content. 
Our effort is similar to this corpus in that it contains content from multiple sign languages from around the world, as well as a large number of videos sourced from YouTube.
One major difference is that, whereas the YouTube-SL-25 videos were selected using an automatic classifier, our videos were manually handpicked.
In addition to what is provided by YouTube-SL-25, our collection includes extensive metadata containing licence information and domain labels. We have also communicated with the content creators to inform them about the data collection and obtain additional permissions.
Finally, we can confirm that the YouTube videos referred to by our collection do not overlap with those included in YouTube-SL-25.


One of the most used sign languages, \textbf{American Sign Language (ASL)}, has the largest amount of available continuous sign language corpora. 
The largest such corpus is YouTube-ASL \cite{tanzerzhang-2024-youtubesl25}, providing 1,000 hours of signed content, sourced from YouTube, as a predecessor of YouTube-SL-25. 
The next largest dataset of considerable size is How2Sign \cite{HOW2SIGN2021}, comprising a vocabulary of 16k tokens and 79 hours of video.
It features 11 signers and provides loosely aligned sentence-level transcriptions, gloss-level transcriptions, OpenPose keypoints, depth camera images, and speech tracks from the video sources.
The FLEURS-ASL~\cite{tanzer-2025-fleursasl} corpus comprises approximately 15 hours of sign language video data and is primarily intended for benchmarking MT models that translate ASL to text of the spoken languages included in FLEURS~\cite{conneau-etal-2023-fleurs} and FLORES~\cite{goyal-etal-2022-flores101}.

\textbf{German Sign Language} (DGS) has the following available continuous sign language corpora. 
The DGS-Corpus\cite{DGS2020} consists of 50 hours of original content expressed in DGS. 
The corpus has been collected by 330 participants across Germany and is intended for linguistic usage. 
RWTH-PHOENIX-Weather \cite{PHOIENIX2012} has been used as a benchmark for machine translation from sign to text, despite having received criticism for its quality \cite{desisto-etal-2022-challenges}.
The videos were derived from the weather forecast of the German public TV broadcasting service PHOENIX and have a resolution of 210x260 pixels, accumulating to 3.25 hours of content. 
DGS-Fabeln-1 \cite{nunnari-etal-2024-dgsfabeln1} is a parallel corpus of German text and videos containing German fairy tales interpreted into DGS by a native DGS signer. 
The corpus is filmed from 7 angles and contains 573 segments of videos with a total duration of 1 hour and 32 minutes, corresponding to 1,428 written sentences.
The size of the DGS parallel corpora that are included in our collection is ten times the size of the DGS corpora available so far. 

For \textbf{French Sign Language} (LSF), Matignon-LSF \cite{halbout-etal-2024-matignonlsf} is a 39-hour corpus of interpreted government speeches, featuring aligned LSF videos with French audio/subtitles.
Dicta-Sign-LSF-v2~\cite{belissen-etal-2020-dictasignlsfv2a} contains 11-hour travel-themed dialogues with 18 signers with annotation on both lexical and non-lexical signs.
MEDIAPI-SKEL~\cite{bull-etal-2020-mediapi} contains 27 hours of video of body, face and hand keypoints of originally-signed sign language, aligned to subtitles.
Rosetta-LSF~\cite{bertin-lemee-etal-2022-rosettalsf} focuses on text-to-sign translation through virtual signers and aligns French news headlines with LSF videos and AZee formal representations.
For \textbf{Spanish Sign Language} (LSE), the corpus that stands out is CORLSE~\cite{deespanola-2023-desafio} with over 380 hours of continuous LSE video recordings, featuring naturalistic and semi-structured discourse from more than 200 deaf signers from diverse regions across Spain.
The KSL dataset \cite{hong-etal-2018-development} is the biggest representant of \textbf{Korean Sign Language} and contains 90 hours of 3-camera recordings of discussions, stories, and lexical elicitation. 

For \textbf{Latin American sign languages} included in our collection, to the best of our knowledge, there is no consistent parallel corpus of continuous sign language. 
Small corpora for isolated signs or focused linguistic studies are available for the sign languages of Argentina~\cite{lsa64-2016,massonecuriel-2004-sign}, Chile~\cite{moralesacostalattapiatnavarro-2024-translation}, Colombia~\cite{felipeflorez-sierra-etal-2024-lsc50}, Costa Rica~\cite{naranjo-zeledon-etal-2020-phonological} and Mexico~\cite{trujillo-romerogarcia-bautista-2023-mexican}. 
To the best of our knowledge, no resources exist for the sign languages of Ecuador, Nicaragua and Peru. 

\section{Construction of the Corpus Collection}
\label{sec:construction}

The construction of the corpus collection consisted of the manual search for the sources, the collection of the licensing information and the processing of the content. 

\subsection{Manual Search of Sign Language Sources}

The sign language content used for this corpus was manually collected by eight hearing individuals, all native speakers of the languages they were responsible for.
This group included native speakers of French, German, Korean and Spanish from Latin America.
Having speakers of the spoken language conduct the search process was crucial, as they could more easily allocate and identify signed content relating to politics, academia or the news, or specific types of videos, using keywords (e.g. 'Cadena' for Spanish, meaning 'National Transmission').
Their familiarity with linguistic and cultural context allowed for more targeted and effective search queries.

It is important to note that, although some countries share the same spoken language, such as the United States of America and United Kingdom with English, for example, the SLs used in each country differ. For instance, ASL is used in the United States of America, while BSL (British Sign Language) is used in the United Kingdom. The same applies to Spanish-speaking countries such as Spain and Latin America. Each country in Latin America has its own official sign language: For example, LSM is used in Mexico, LSA in Argentina and LSCh in Chile. The same pattern applies to the SLs of Arabic- and Portuguese-speaking countries.

A significant portion of the search was conducted on the media platform YouTube, which has an interface that allows users to search for videos by topic or channel and apply filters such as the presence of subtitles or closed captions, and whether they are licensed under Creative Commons.
This was helpful for clarifying licence requirements, which will be explained later.
Nowadays, it is common for news and government channels to include a sign language interpreter, so these were also searched for.
These videos either show the interpreter in one of the lower corners of the screen or alongside the person speaking.
However, not all videos from news or government channels include a sign language interpreter, so an extensive search was necessary.
In some cases, we created custom YouTube playlists to isolate sign language videos from channels with a wider range of content.

\subsection{Collecting Licensing Information}

The vision behind creating this collection was to make the corpora as freely and publicly available as possible.
Therefore, as previously mentioned, the initial search focused on online sources that publish content under permissive licences.
Secondly, we verified the licences of the videos as they were collected individually.
Finally, we communicated with the content creators, providing them with a detailed description of our project and its purposes.
This enabled us to obtain informed consent for the data collection and additionally resulted in positive permissions being granted, which were not initially available.

The collected corpora are covered by the licences of \textit{public domain}, \textit{creative commons}, \textit{YouTube standard license}, \textit{permission for research usage}, and \textit{permission for research usage after keypoint extraction}. 

\begin{table}
\small
\begin{tabularx}{0.45\textwidth}{Xl}
    \toprule

    \thead{Name} & \thead{Unit/format} \\

    \midrule

    \rowcolor{rowA} ID & - \\
    \rowcolor{rowB} Channel ID & - \\    
    \rowcolor{rowA} Video name & - \\
    \rowcolor{rowB} Release date & YYYY-MM-DD \\
    \rowcolor{rowA} Website URL & - \\
    \rowcolor{rowB} Video URL & - \\
    \rowcolor{rowA} Resolution & pixels x pixels \\
    \rowcolor{rowB} Video length & HH:MM:SS \\
    \rowcolor{rowA} Frequency & FPS \\
    \rowcolor{rowB} Date of acquisition & YYYY-MM-DD \\    
    \rowcolor{rowA} File size & MegaBytes \\
    \rowcolor{rowB} Subtitles & -  \\    
    \rowcolor{rowA} Sentences & - \\
    \rowcolor{rowB} Tokens & - \\    
    \rowcolor{rowA} Characters & - \\
    
    \bottomrule
    
\end{tabularx}
\caption{Metadata table}
\label{table:metadata}
\end{table}

\subsection{Processing of content}
Subtitles had to be extracted from videos in order to calculate spoken language statistics and include them in the metadata.
In some cases, optical character recognition (OCR) was used to extract subtitles that were part of the video.
A sentence and word tokeniser was used to tokenise the subtitle text.

Videos were often cropped to isolate the interpreter when they were blended into part of the screen during spoken shows.
Cropping the interpreter enables the resolution of the signed content to be calculated, which is a useful technical quality indicator.

\begin{table*}
\centering
\begin{tabular}{llrrrrrrrr}
\toprule
Spoken Language/Dialect & Sign Language & Video Length & File Size (GB) & Videos & Subtitles & Tokens & Characters \\
\midrule
Argentinian     & LSA           & 78:54:31      & 26             & 306    & 80,480     & 456,910 & 2,555,712    \\
Chilean         & LSCh          & 94:48:59      & 41             & 379    & 71,047     & 411,961 & 2,316,400    \\
Colombian       & LSC           & 118:15:47     & 51             & 827    & 143,973    & 834,068 & 4,768,925    \\
Costa Rican     & LESCO         & 196:57:24     & 56             & 508    & 19,626     & 120,151 & 680,377     \\
Ecuatorian      & LSEC          & 27:43:56      & 5              & 41     & 32,612     & 193,187 & 1,118,958    \\
French          & LSF           & 06:50:20      & 4              & 73     & 8,621      & 58,640  & 298,368     \\
German          & DGS           & 529:14:52   & 149            & 578    & 541,948    & 9,669,699 & 59,706,371   \\
Korean          & KSL           & 04:28:57      & 3              & 44     & 6,322      & 27,052  & 96,818      \\
Mexican         & LSM           & 50:13:52      & 17             & 247    & 55,801     & 281,021 & 1,524,000    \\
Nicaraguan      & ISN           & 05:01:25      & 3              & 30     & 4,442      & 15,021  & 82,576      \\
Peruvian        & LSP           & 270:27:43     & 81             & 1,215   & 362,766    & 2,195,060 & 12,502,837   \\
Spanish (*Castilian)         & LSE           & 08:17:15      & 5              & 133    & 7,682      & 42,867  & 235,177     \\
\midrule
\textbf{Sum}    &               & 1,391:25:01    & 440         & 4,381   & 1,335,320   & 14,305,637 & 85,886,519   \\
\bottomrule
\end{tabular}
\caption{Corpora statistics aggregated by language pair}
\label{tab:language_statistics}
\end{table*}

\section{Description of the Corpora}
\label{sec:corpora_description}

Here we will provide details about the metadata catalogue, statistics on the data and content characteristics of the videos. 

\subsection{Catalogue of Metadata}
Following previous work, and due to redistribution restrictions on a substantial portion of the collection, we provide the URL addresses together with extensive metadata. 
The catalogue of the metadata is provided in two levels:

\paragraph{Channel list.} 
The first metadata CSV table lists the \textit{channels} (i.e. content sources), with one row per channel.
Aggregated metadata are provided for each channel, including information on the sign language covered, the licence, the total length and size of videos, the number of subtitles, and the number of sentences, tokens and characters they contain. The time period during which the original material was released is also provided.

\paragraph{Video list.} 
The videos are listed in a CSV table for each collection. There is one row for each file, which is identified by an incremental ID and accompanied by the relevant video file metadata (\Cref{table:metadata}).
The first part of the metadata provides a description of the video's basic characteristics, including the video name, website and video URLs, and the release date.
This is followed by technical details such as resolution, video length, frequency in frames per second (FPS), date of acquisition, and file size.
The final section refers to the associated content in the relevant spoken language (if available), as well as the number of subtitles, sentences, tokens and characters.

\subsection{Data overview}
An overview of the size of the corpora for every language is provided in \Cref{tab:language_statistics}. 
The corpus collection consists of 4,381 videos in 12 sign languages.
The most represented spoken-sign language pairs in terms of video duration are those from Germany, Peru, Costa Rica and Colombia.

The second overview in \Cref{tab:channel_statistics} provides details on the content sources that have been included in the corpus for each language. 
It is also possible to view additional details here, such as the domain to which each content source belongs.
\begin{table*}
\footnotesize
\begin{tabular}{l l p{2.4cm} l l r r r r r r l l}
\toprule
SL & Spoken L & Name & Content & Licence & Videos & Length & (GB) & Subtitles & Tokens & Characters & From & To \\
\midrule
LSA & Argentinian & Aprender entre rios & Edu & CC & 121 & 05:55:53 & 3 & 6,316 & 36,398 & 201,226 & 2015-07-31 & 2022-10-05 \\
LSA & Argentinian & Universidad Nacional de Entre Ríos - Canal 20 & Edu & CC & 69 & 05:34:02 & 4 &    &    &    & 2014-01-10 & 2022-10-05 \\
LSA & Argentinian & Casa Rosada - República Argentina & Gov & Public & 116 & 67:24:36 & 19 & 74,164 & 420,512 & 2,354,486 & 2010-12-20 & 2023-12-01 \\
\midrule
LSCh & Chilean & SaludResponde & Health & YT & 17 & 00:12:45 &    & 256 & 1,503 & 8,469 & 2020-04-08 & 2022-07-20 \\
LSCh & Chilean & Lense Biobio Chile & Edu & YT & 36 & 14:19:59 & 4 & 13,262 & 66,151 & 344,427 & 2020-03-26 & 2022-08-08 \\
LSCh & Chilean & BCNChile & Gov & CC & 20 & 00:27:04 &    & 523 & 2,872 & 17,520 & 2021-03-19 & 2022-04-04 \\
LSCh & Chilean & BCNChile & Gov & CC & 203 & 34:27:45 & 17 &    &    &    & 2013-05-09 & 2022-10-13 \\
LSCh & Chilean & Prensa Presidencia & Gov & CC & 7 & 00:48:05 &    & 1,218 & 7,139 & 42,676 & 2015-12-31 & 2017-12-31 \\
LSCh & Chilean & Cauquenesnet La Web de Cauquenes & News & CC & 12 & 01:26:27 & 1 & 1,623 & 9,003 & 47,814 & 2014-01-12 & 2023-08-01 \\
LSCh & Chilean & Timeline Antofagasta & News & CC & 20 & 11:05:29 & 7 & 13,547 & 79,802 & 454,352 & 2021-04-25 & 2022-12-10 \\
LSCh & Chilean & Terrenos de Chile & Gov & CC & 64 & 32:01:25 & 11 & 40,618 & 245,491 & 1,401,142 & 2019-09-27 & 2021-12-19 \\
\midrule
LSC & Colombian & Fundesor & Social & YT & 13 & 00:22:58 &    &    &    &    & 2015-07-23 & 2016-09-27 \\
LSC & Colombian & FENASCOL & Social & CC & 617 & 87:51:25 & 33 & 105,280 & 610,678 & 3,472,239 & 2011-05-08 & 2023-05-01 \\
LSC & Colombian & Ayudas Para Todos & Edu & YT & 12 & 00:52:55 &    & 865 & 4,363 & 23,569 & 2013-07-16 & 2020-09-23 \\
LSC & Colombian & Ministerio TIC Colombia & Gov & CC & 50 & 04:47:24 & 3 & 5,951 & 34,627 & 198,409 & 2013-05-11 & 2021-08-24 \\
LSC & Colombian & Presidencia de la República - Colombia & Gov & YT & 135 & 24:21:05 & 15 & 31,877 & 184,400 & 1,074,708 & 2013-02-10 & 2022-12-23 \\
\midrule
LESCO & Costa rican & Presidencia de la República & Gov & YT & 440 & 182:45:58 & 53 &    &    &    & 2018-05-23 & 2022-12-31 \\
LESCO & Costa rican & OndaUNED & News & CC & 27 & 12:07:10 & 2 & 17,038 & 104,322 & 592,683 & 2014-09-25 & 2022-05-15 \\
LESCO & Costa rican & La Reaccion & News & YT & 41 & 02:04:16 & 1 & 2,588 & 15,829 & 87,694 & 2021-01-11 & 2022-11-08 \\
\midrule
LSEC & Ecuatorian & TELE CIUDADANA & News & CC & 41 & 27:43:56 & 5 & 32,612 & 193,187 & 1,118,958 & 2015-05-19 & 2016-10-02 \\
\midrule
LSF & French & Littlebunbao & Edu & YT & 73 & 06:50:20 & 4 & 8,621 & 58,640 & 298,368 & 2018-01-16 & 2022-08-21 \\
\midrule
KSL & Korean & \begin{CJK}{UTF8}{mj} 말해주세요 \#새정부에바란다 \end{CJK}& News & Public & 25 & 02:17:49 & 1 & 3,249 & 12,747 & 47,052 & & \\
KSL & Korean & \begin{CJK}{UTF8}{mj} 뉴텔러 - 키워드로 새롭게 보는 정책 \end{CJK} & Gov & Public & 19 & 02:11:08 & 1 & 3,073 & 14,305 & 49,766 & & \\
\midrule
DGS & German & Bundestag & Gov & 3Keypoints & 388 & 525:00:43 & 132 & 454,700 & 9,065,020 & 55,548,592 \\
 DGS  &  German & Heute Journal & News & Research & 190 & 04:29:04 & 18 & 87,248 & 604,679 & 4,157,779 \\
\midrule
LSM & Mexican & Andrés Manuel López Obrador & Gov & YT & 154 & 31:47:31 & 7 & 34,876 & 165,170 & 916,187 & 2020-07-28 & 2023-06-01 \\
LSM & Mexican & LSM IAPPPS & Edu & YT & 33 & 04:28:13 & 3 & 5,656 & 33,984 & 189,680 & 2020-01-04 & 2022-10-31 \\
LSM & Mexican & LSM Enseñando & Edu & YT & 41 & 09:04:25 & 3 & 9,599 & 44,873 & 233,340 & 2020-01-07 & 2022-10-31 \\
LSM & Mexican & CuriosaMente & Edu & YT & 1 & 00:08:38 &    &    &    &    & 2021-02-05 & 2021-02-05 \\
LSM & Mexican & Brenda Mtz A & Vlog & YT & 18 & 04:45:05 & 3 & 5,670 & 36,994 & 184,793 & 2021-12-14 & 2022-11-02 \\
\midrule
ISN & Nicaraguan & VOS TV & Edu & CC & 30 & 05:01:25 & 3 & 4,442 & 15,021 & 82,576 & 2020-01-10 & 2021-12-01 \\
\midrule
LSP & Peruvian & Defensoría del Pueblo Perú & Gov & YT & 1,215 & 270:27:43 & 81 & 362,766 & 2,195,060 & 12,502,837 & 2011-01-04 & 2023-06-01 \\
\midrule
LSE & Spanish* & CÓRDOBA HOY & News & YT & 34 & 04:04:45 & 3 & 4,987 & 28,675 & 160,174 & 2021-10-10 & 2022-12-06 \\
LSE & Spanish* & MariaOrtiz-LSEenfermeria & Edu & YT & 52 & 00:56:24 &    & 727 & 1,948 & 12,170 & & \\
LSE & Spanish* & Aprender Gratis & Edu & YT & 25 & 01:14:47 & 1 & 216 & 1,237 & 6,153 & 2020-09-17 & 2022-12-26 \\
LSE & Spanish* & otanana & Edu & CC & 22 & 02:01:19 & 1 & 1,752 & 11,007 & 56,680 & 2013-07-26 & 2022-12-18 \\
\bottomrule
\end{tabular}
\caption{Details for the content sources (channels) included in the collection. YT: YouTube Standard License}
\label{tab:channel_statistics}
\end{table*}

\subsection{Content types}

The content sources are predominantly labelled with one of four content types:

\paragraph{Political Content}
In terms of video length, political content such as press conferences, parliamentary sessions and official government communications constitutes the largest category in the corpus.
Videos from the German parliament (Bundestag) fall under this category, containing the majority of content in DGS. There are 388 videos in this category, with a combined length of 525 hours and half a million transcribed sentences.
This category also includes presidential speeches and government content from several Latin American countries and South Korea.

\paragraph{News Broadcasts}
The news category originates from television broadcasts and includes, among others, the German news bulletin `Heute Journal' (with an average length of 20 minutes), the Chilean news outlet `Timeline Antofagasta', and the Ecuadorian Executive Branch Official Channel `Tele Ciudadana'.

\paragraph{Educational Content}
This category comprises various sources related to educational contexts.
These include educational institutions, such as universities, or organisations and channels that promote sign language learning.
Unlike the majority of videos in the corpus, which show interpreted content, this category also contains material that is originally produced in a sign language.

\paragraph{Social Content}
This content refers to videos derived from social organizations, such as unions of the deaf and NGOs.

\subsection{Subtitles}
One of the core criteria for selecting videos is the availability of subtitles.
Subtitles are attached to videos linked by our metadata lists.
They are typically provided in VTT, SRT and XML formats (the latter of which allows for word-level segmentation of broader subtitle units).

There are two types of subtitles: those that have been manually created by transcribers and provided by the content providers, and those that have been auto-generated by the video platform.
Manually created subtitles are undoubtedly of a higher quality than automatically generated subtitles, which are created using automatic speech recognition and may contain errors.
Nevertheless, we believe that auto-generated subtitles are better than no subtitles, and therefore we include them in the collection.
However, their usability for relevant tasks must be assessed.
Information on whether the subtitles are manually or automatically generated is provided in the metadata of every channel.
For auto-generated subtitles, the number of sentences is not reported due to a lack of punctuation.

\section{Limitations and Further Work}
\label{sec:limitations}

Although we performed careful manual selection and metadata annotation of the sign language videos, the vast majority of the process has been done by hearing persons, native speakers of the spoken languages on the textual side of the parallel corpora. 
Despite our efforts, we could only include one signer of DGS but no signers of the other languages in the project. 
This may entail issues for the identification of the sign languages, the quality of the closed captions, but also requires future circumspection about the compliance with FAIR and CARE principles \cite{carroll-etal-2020-care, wilkinson-etal-2016-fair} with regard to the local signing communities. 
Efforts should be made in further work, and as we did not have access to local communities during the development of the collection, we kindly invite them to engage with the open source content towards its improvement. 

As in most interpreted content, the subtitles provided are a transcription of the spoken content and not the sign language. 
As the subtitles and the sign language interpretation have been produced most probably independent of each other, based on the spoken content, there may be considerable deviations between the two. 
An isolated portion of the screen which contains the signing person is not provided in the current version due to licence limitations.
We aim to provide co-ordinates and scripts in future updates. 

An additional challenge is the timely alignment of the subtitles with the signed content, due to the inevitable variable delay introduced during interpretation. 
This issue may need to be addressed with further technical means such as segmentation~\cite{moryossef-etal-2023-linguistically, renz-etal-2021-sign} or automatic shifting of the subtitles to match the signed content~\cite{bull-etal-2021-aligning}. 
We intend to address this limitation in further versions of our corpus.

\section{Conclusion}
\label{sec:conclusion}
The TUB Sign Language Corpus Collection represents a significant step toward the development of multilingual sign language resources by assembling a large-scale, diverse, and richly annotated dataset of signed videos and corresponding spoken language subtitles. 
With over 1,300 hours of video across 12 sign languages and extensive metadata, the collection aims to provide a valuable foundation for further research on sign language technology. 
Future work will focus on increasing the data and their availability, improving temporal alignment, and enhancing community engagement, to better serve both scientific and signing communities.

\begin{acks}
The work reported in this paper was conducted as part of the module ``Interdisciplinary Media Project'' at the Quality and Usability Lab of the Technical University of Berlin. It was supported by BMBF (German Federal Ministry of Education and Research) via the project SocialWear (grant no.~01IW20002) and by the European Union via the project SignReality, as part of financial support to third parties by the UTTER project (Horizon Europe, GA: 101070631). Thanks to Laura Schowe and Ahmed Boulila for their participation in the initial corpus collection efforts. 
\end{acks}

\bibliographystyle{ACM-Reference-Format}
\bibliography{bib1, corpora}


\begin{thebibliography}{26}


\ifx \showCODEN    \undefined \def \showCODEN     #1{\unskip}     \fi
\ifx \showISBNx    \undefined \def \showISBNx     #1{\unskip}     \fi
\ifx \showISBNxiii \undefined \def \showISBNxiii  #1{\unskip}     \fi
\ifx \showISSN     \undefined \def \showISSN      #1{\unskip}     \fi
\ifx \showLCCN     \undefined \def \showLCCN      #1{\unskip}     \fi
\ifx \shownote     \undefined \def \shownote      #1{#1}          \fi
\ifx \showarticletitle \undefined \def \showarticletitle #1{#1}   \fi
\ifx \showURL      \undefined \def \showURL       {\relax}        \fi
\providecommand\bibfield[2]{#2}
\providecommand\bibinfo[2]{#2}
\providecommand\natexlab[1]{#1}
\providecommand\showeprint[2][]{arXiv:#2}

\bibitem[Belissen et~al\mbox{.}(2020)]%
        {belissen-etal-2020-dictasignlsfv2a}
\bibfield{author}{\bibinfo{person}{Valentin Belissen},
  \bibinfo{person}{Annelies Braffort}, {and} \bibinfo{person}{Mich{\`e}le
  Gouiff{\`e}s}.} \bibinfo{year}{2020}\natexlab{}.
\newblock \showarticletitle{Dicta-{{Sign-LSF-v2}}: {{Remake}} of a {{Continuous
  French Sign Language Dialogue Corpus}} and a {{First Baseline}} for
  {{Automatic Sign Language Processing}}}. In
  \bibinfo{booktitle}{\emph{Proceedings of the {{Twelfth Language Resources}}
  and {{Evaluation Conference}}}}, \bibfield{editor}{\bibinfo{person}{Nicoletta
  Calzolari}, \bibinfo{person}{Fr{\'e}d{\'e}ric B{\'e}chet},
  \bibinfo{person}{Philippe Blache}, \bibinfo{person}{Khalid Choukri},
  \bibinfo{person}{Christopher Cieri}, \bibinfo{person}{Thierry Declerck},
  \bibinfo{person}{Sara Goggi}, \bibinfo{person}{Hitoshi Isahara},
  \bibinfo{person}{Bente Maegaard}, \bibinfo{person}{Joseph Mariani},
  \bibinfo{person}{H{\'e}l{\`e}ne Mazo}, \bibinfo{person}{Asuncion Moreno},
  \bibinfo{person}{Jan Odijk}, {and} \bibinfo{person}{Stelios Piperidis}}
  (Eds.). \bibinfo{publisher}{European Language Resources Association},
  \bibinfo{address}{Marseille, France}, \bibinfo{pages}{6040--6048}.
\newblock
\showISBNx{979-10-95546-34-4}
\urldef\tempurl%
\url{https://aclanthology.org/2020.lrec-1.740/}
\showURL{%
\tempurl}


\bibitem[{Bertin-Lem{\'e}e} et~al\mbox{.}(2022)]%
        {bertin-lemee-etal-2022-rosettalsf}
\bibfield{author}{\bibinfo{person}{Elise {Bertin-Lem{\'e}e}},
  \bibinfo{person}{Annelies Braffort}, \bibinfo{person}{Camille Challant},
  \bibinfo{person}{Claire Danet}, \bibinfo{person}{Boris Dauriac},
  \bibinfo{person}{Michael Filhol}, \bibinfo{person}{Emmanuella Martinod},
  {and} \bibinfo{person}{J{\'e}r{\'e}mie Segouat}.}
  \bibinfo{year}{2022}\natexlab{}.
\newblock \showarticletitle{Rosetta-{{LSF}}: An {{Aligned Corpus}} of {{French
  Sign Language}} and {{French}} for {{Text-to-Sign Translation}}}. In
  \bibinfo{booktitle}{\emph{Proceedings of the {{Thirteenth Language
  Resources}} and {{Evaluation Conference}}}},
  \bibfield{editor}{\bibinfo{person}{Nicoletta Calzolari},
  \bibinfo{person}{Fr{\'e}d{\'e}ric B{\'e}chet}, \bibinfo{person}{Philippe
  Blache}, \bibinfo{person}{Khalid Choukri}, \bibinfo{person}{Christopher
  Cieri}, \bibinfo{person}{Thierry Declerck}, \bibinfo{person}{Sara Goggi},
  \bibinfo{person}{Hitoshi Isahara}, \bibinfo{person}{Bente Maegaard},
  \bibinfo{person}{Joseph Mariani}, \bibinfo{person}{H{\'e}l{\`e}ne Mazo},
  \bibinfo{person}{Jan Odijk}, {and} \bibinfo{person}{Stelios Piperidis}}
  (Eds.). \bibinfo{publisher}{European Language Resources Association},
  \bibinfo{address}{Marseille, France}, \bibinfo{pages}{4955--4962}.
\newblock
\urldef\tempurl%
\url{https://aclanthology.org/2022.lrec-1.529/}
\showURL{%
\tempurl}


\bibitem[Bull et~al\mbox{.}(2021)]%
        {bull-etal-2021-aligning}
\bibfield{author}{\bibinfo{person}{Hannah Bull}, \bibinfo{person}{Triantafyllos
  Afouras}, \bibinfo{person}{G{\"u}l Varol}, \bibinfo{person}{Samuel Albanie},
  \bibinfo{person}{Liliane Momeni}, {and} \bibinfo{person}{Andrew Zisserman}.}
  \bibinfo{year}{2021}\natexlab{}.
\newblock \showarticletitle{Aligning {{Subtitles}} in {{Sign Language
  Videos}}}.
\newblock \bibinfo{journal}{\emph{Proceedings of the IEEE International
  Conference on Computer Vision}} (\bibinfo{date}{May} \bibinfo{year}{2021}),
  \bibinfo{pages}{11532--11541}.
\newblock
\showISBNx{9781665428125}
\showISSN{15505499}
\href{https://doi.org/10.1109/ICCV48922.2021.01135}{doi:\nolinkurl{10.1109/ICCV48922.2021.01135}}
\showeprint[arxiv]{2105.02877}


\bibitem[Bull et~al\mbox{.}(2020)]%
        {bull-etal-2020-mediapi}
\bibfield{author}{\bibinfo{person}{Hannah Bull}, \bibinfo{person}{Annelies
  Braffort}, {and} \bibinfo{person}{Mich{\`e}le Gouiff{\`e}s}.}
  \bibinfo{year}{2020}\natexlab{}.
\newblock \showarticletitle{{MEDIAPI}-{SKEL} - A 2{D}-Skeleton Video Database
  of {F}rench {S}ign {L}anguage With Aligned {F}rench Subtitles}. In
  \bibinfo{booktitle}{\emph{Proceedings of the Twelfth Language Resources and
  Evaluation Conference}}, \bibfield{editor}{\bibinfo{person}{Nicoletta
  Calzolari}, \bibinfo{person}{Fr{\'e}d{\'e}ric B{\'e}chet},
  \bibinfo{person}{Philippe Blache}, \bibinfo{person}{Khalid Choukri},
  \bibinfo{person}{Christopher Cieri}, \bibinfo{person}{Thierry Declerck},
  \bibinfo{person}{Sara Goggi}, \bibinfo{person}{Hitoshi Isahara},
  \bibinfo{person}{Bente Maegaard}, \bibinfo{person}{Joseph Mariani},
  \bibinfo{person}{H{\'e}l{\`e}ne Mazo}, \bibinfo{person}{Asuncion Moreno},
  \bibinfo{person}{Jan Odijk}, {and} \bibinfo{person}{Stelios Piperidis}}
  (Eds.). \bibinfo{publisher}{European Language Resources Association},
  \bibinfo{address}{Marseille, France}, \bibinfo{pages}{6063--6068}.
\newblock
\showISBNx{979-10-95546-34-4}
\urldef\tempurl%
\url{https://aclanthology.org/2020.lrec-1.743/}
\showURL{%
\tempurl}


\bibitem[Carroll et~al\mbox{.}(2020)]%
        {carroll-etal-2020-care}
\bibfield{author}{\bibinfo{person}{Stephanie~Russo Carroll},
  \bibinfo{person}{Ibrahim Garba}, \bibinfo{person}{Oscar~L.
  {Figueroa-Rodr{\'i}guez}}, \bibinfo{person}{Jarita Holbrook},
  \bibinfo{person}{Raymond Lovett}, \bibinfo{person}{Simeon Materechera},
  \bibinfo{person}{Mark Parsons}, \bibinfo{person}{Kay Raseroka},
  \bibinfo{person}{Desi {Rodriguez-Lonebear}}, \bibinfo{person}{Robyn Rowe},
  \bibinfo{person}{Rodrigo Sara}, \bibinfo{person}{Jennifer~D. Walker},
  \bibinfo{person}{Jane Anderson}, {and} \bibinfo{person}{Maui Hudson}.}
  \bibinfo{year}{2020}\natexlab{}.
\newblock \showarticletitle{The {{CARE Principles}} for {{Indigenous Data
  Governance}}}.
\newblock \bibinfo{journal}{\emph{Data Science Journal}}  \bibinfo{volume}{19}
  (\bibinfo{date}{Nov.} \bibinfo{year}{2020}), \bibinfo{pages}{43--43}.
\newblock
\showISSN{1683-1470}
\href{https://doi.org/10.5334/dsj-2020-043}{doi:\nolinkurl{10.5334/dsj-2020-043}}


\bibitem[Conneau et~al\mbox{.}(2023)]%
        {conneau-etal-2023-fleurs}
\bibfield{author}{\bibinfo{person}{Alexis Conneau}, \bibinfo{person}{Min Ma},
  \bibinfo{person}{Simran Khanuja}, \bibinfo{person}{Yu Zhang},
  \bibinfo{person}{Vera Axelrod}, \bibinfo{person}{Siddharth Dalmia},
  \bibinfo{person}{Jason Riesa}, \bibinfo{person}{Clara Rivera}, {and}
  \bibinfo{person}{Ankur Bapna}.} \bibinfo{year}{2023}\natexlab{}.
\newblock \showarticletitle{{{FLEURS}}: {{FEW-Shot Learning Evaluation}} of
  {{Universal Representations}} of {{Speech}}}. In
  \bibinfo{booktitle}{\emph{2022 {{IEEE Spoken Language Technology Workshop}}
  ({{SLT}})}}. \bibinfo{pages}{798--805}.
\newblock
\href{https://doi.org/10.1109/SLT54892.2023.10023141}{doi:\nolinkurl{10.1109/SLT54892.2023.10023141}}


\bibitem[De~Sisto et~al\mbox{.}(2022)]%
        {desisto-etal-2022-challenges}
\bibfield{author}{\bibinfo{person}{Mirella De~Sisto}, \bibinfo{person}{Vincent
  Vandeghinste}, \bibinfo{person}{Santiago Egea~G{\'o}mez},
  \bibinfo{person}{Mathieu De~Coster}, \bibinfo{person}{Dimitar Shterionov},
  {and} \bibinfo{person}{Horacio Saggion}.} \bibinfo{year}{2022}\natexlab{}.
\newblock \showarticletitle{Challenges with {{Sign Language Datasets}} for
  {{Sign Language Recognition}} and {{Translation}}}. In
  \bibinfo{booktitle}{\emph{Proceedings of the {{Thirteenth Language
  Resources}} and {{Evaluation Conference}}}}. \bibinfo{publisher}{European
  Language Resources Association}, \bibinfo{address}{Marseille, France},
  \bibinfo{pages}{2478--2487}.
\newblock
\urldef\tempurl%
\url{https://aclanthology.org/2022.lrec-1.264}
\showURL{%
\tempurl}


\bibitem[Duarte et~al\mbox{.}(2021)]%
        {HOW2SIGN2021}
\bibfield{author}{\bibinfo{person}{Amanda Duarte}, \bibinfo{person}{Shruti
  Palaskar}, \bibinfo{person}{Lucas Ventura}, \bibinfo{person}{Deepti
  Ghadiyaram}, \bibinfo{person}{Kenneth DeHaan}, \bibinfo{person}{Florian
  Metze}, \bibinfo{person}{Jordi Torres}, {and} \bibinfo{person}{Xavier Giro-i
  Nieto}.} \bibinfo{year}{2021}\natexlab{}.
\newblock \showarticletitle{How2sign: a large-scale multimodal dataset for
  continuous american sign language}. In \bibinfo{booktitle}{\emph{Proceedings
  of the IEEE/CVF conference on computer vision and pattern recognition}}.
  \bibinfo{pages}{2735--2744}.
\newblock


\bibitem[{Felipe Fl{\'o}rez-Sierra} et~al\mbox{.}(2024)]%
        {felipeflorez-sierra-etal-2024-lsc50}
\bibfield{author}{\bibinfo{person}{Andr{\'e}s {Felipe Fl{\'o}rez-Sierra}},
  \bibinfo{person}{Brayan David~Sol{\'o}rzano}, \bibinfo{person}{Fredy
  {Segura-Quijano}}, \bibinfo{person}{Yenny~Milena {Cort{\'e}s-Bello}},
  \bibinfo{person}{Luis Cubillos}, \bibinfo{person}{Luis~Felipe Giraldo}, {and}
  \bibinfo{person}{Christian {Cifuentes-De la Portilla}}.}
  \bibinfo{year}{2024}\natexlab{}.
\newblock \showarticletitle{{{LSC50}}: {{Colombian Sign Language Video}} and
  {{Inertial Measurement}} Dataset}.
\newblock \bibinfo{journal}{\emph{Scientific Data}} \bibinfo{volume}{11},
  \bibinfo{number}{1} (\bibinfo{date}{Dec.} \bibinfo{year}{2024}),
  \bibinfo{pages}{1347}.
\newblock
\showISSN{2052-4463}
\href{https://doi.org/10.1038/s41597-024-04172-5}{doi:\nolinkurl{10.1038/s41597-024-04172-5}}


\bibitem[Forster et~al\mbox{.}(2012)]%
        {PHOIENIX2012}
\bibfield{author}{\bibinfo{person}{Jens Forster}, \bibinfo{person}{Christoph
  Schmidt}, \bibinfo{person}{Thomas Hoyoux}, \bibinfo{person}{Oscar Koller},
  \bibinfo{person}{Uwe Zelle}, \bibinfo{person}{Justus~H Piater}, {and}
  \bibinfo{person}{Hermann Ney}.} \bibinfo{year}{2012}\natexlab{}.
\newblock \showarticletitle{RWTH-PHOENIX-weather: A large vocabulary sign
  language recognition and translation corpus.}. In
  \bibinfo{booktitle}{\emph{LREC}}, Vol.~\bibinfo{volume}{9}.
  \bibinfo{pages}{3785--3789}.
\newblock


\bibitem[Goyal et~al\mbox{.}(2022)]%
        {goyal-etal-2022-flores101}
\bibfield{author}{\bibinfo{person}{Naman Goyal}, \bibinfo{person}{Cynthia Gao},
  \bibinfo{person}{Vishrav Chaudhary}, \bibinfo{person}{Peng-Jen Chen},
  \bibinfo{person}{Guillaume Wenzek}, \bibinfo{person}{Da Ju},
  \bibinfo{person}{Sanjana Krishnan}, \bibinfo{person}{Marc'Aurelio Ranzato},
  \bibinfo{person}{Francisco Guzm{\'a}n}, {and} \bibinfo{person}{Angela Fan}.}
  \bibinfo{year}{2022}\natexlab{}.
\newblock \showarticletitle{The {{Flores-101 Evaluation Benchmark}} for
  {{Low-Resource}} and {{Multilingual Machine Translation}}}.
\newblock \bibinfo{journal}{\emph{Transactions of the Association for
  Computational Linguistics}}  \bibinfo{volume}{10} (\bibinfo{year}{2022}),
  \bibinfo{pages}{522--538}.
\newblock
\href{https://doi.org/10.1162/tacl_a_00474}{doi:\nolinkurl{10.1162/tacl_a_00474}}


\bibitem[Halbout et~al\mbox{.}(2024)]%
        {halbout-etal-2024-matignonlsf}
\bibfield{author}{\bibinfo{person}{Julie Halbout}, \bibinfo{person}{Diandra
  Fabre}, \bibinfo{person}{Yanis Ouakrim}, \bibinfo{person}{Julie Lascar},
  \bibinfo{person}{Annelies Braffort}, \bibinfo{person}{Mich{\`e}le
  Gouiff{\`e}s}, {and} \bibinfo{person}{Denis Beautemps}.}
  \bibinfo{year}{2024}\natexlab{}.
\newblock \showarticletitle{Matignon-{{LSF}}: A {{Large Corpus}} of
  {{Interpreted French Sign Language}}}. In
  \bibinfo{booktitle}{\emph{Proceedings of the {{LREC-COLING}} 2024 11th
  {{Workshop}} on the {{Representation}} and {{Processing}} of {{Sign
  Languages}}: {{Evaluation}} of {{Sign Language Resources}}}},
  \bibfield{editor}{\bibinfo{person}{ELRA Language~Resources
  Association~(ELRA)} {and} \bibinfo{person}{the International Committee
  on~Computational Linguistics~(ICCL)}} (Eds.). \bibinfo{address}{Turin,
  Italy}, \bibinfo{pages}{202--208}.
\newblock
\urldef\tempurl%
\url{https://hal.science/hal-04593865}
\showURL{%
\tempurl}


\bibitem[Hanke et~al\mbox{.}(2020)]%
        {DGS2020}
\bibfield{author}{\bibinfo{person}{Thomas Hanke}, \bibinfo{person}{Marc
  Schulder}, \bibinfo{person}{Reiner Konrad}, {and} \bibinfo{person}{Elena
  Jahn}.} \bibinfo{year}{2020}\natexlab{}.
\newblock \showarticletitle{Extending the Public DGS Corpus in size and depth}.
  In \bibinfo{booktitle}{\emph{sign-lang@ LREC 2020}}. European Language
  Resources Association (ELRA), \bibinfo{pages}{75--82}.
\newblock


\bibitem[Hong et~al\mbox{.}(2018)]%
        {hong-etal-2018-development}
\bibfield{author}{\bibinfo{person}{Sung-Eun Hong}, \bibinfo{person}{Seongok
  Won}, \bibinfo{person}{Il Heo}, {and} \bibinfo{person}{Hyunhwa Lee}.}
  \bibinfo{year}{2018}\natexlab{}.
\newblock \showarticletitle{Development of an ``{{Integrative System}} for
  {{Korean Sign Language Resources}}''}. In
  \bibinfo{booktitle}{\emph{Sign-Lang@ {{LREC}} 2018}}.
  \bibinfo{publisher}{European Language Resources Association (ELRA)},
  \bibinfo{pages}{75--78}.
\newblock
\urldef\tempurl%
\url{http://lrec-conf.org/workshops/lrec2018/W1/pdf/18031_W1.pdf}
\showURL{%
\tempurl}


\bibitem[Massone and Curiel(2004)]%
        {massonecuriel-2004-sign}
\bibfield{author}{\bibinfo{person}{Maria~Ignacia Massone} {and}
  \bibinfo{person}{Monica Curiel}.} \bibinfo{year}{2004}\natexlab{}.
\newblock \showarticletitle{Sign {{Order}} in {{Argentine Sign Language}}}.
\newblock \bibinfo{journal}{\emph{Sign Language Studies}} \bibinfo{volume}{5},
  \bibinfo{number}{1} (\bibinfo{year}{2004}), \bibinfo{pages}{63--93}.
\newblock
\showISSN{1533-6263}
\href{https://doi.org/10.1353/sls.2004.0023}{doi:\nolinkurl{10.1353/sls.2004.0023}}


\bibitem[Morales~Acosta and Lattapiat~Navarro(2024)]%
        {moralesacostalattapiatnavarro-2024-translation}
\bibfield{author}{\bibinfo{person}{Gina~Viviana Morales~Acosta} {and}
  \bibinfo{person}{Pamela Lattapiat~Navarro}.} \bibinfo{year}{2024}\natexlab{}.
\newblock \showarticletitle{Translation of a Story into {{Chilean Sign
  Language}}: Productions by Deaf Co-Teachers}.
\newblock \bibinfo{journal}{\emph{Frontiers in Education}}  \bibinfo{volume}{9}
  (\bibinfo{date}{Oct.} \bibinfo{year}{2024}).
\newblock
\showISSN{2504-284X}
\href{https://doi.org/10.3389/feduc.2024.1368082}{doi:\nolinkurl{10.3389/feduc.2024.1368082}}


\bibitem[Moryossef et~al\mbox{.}(2023)]%
        {moryossef-etal-2023-linguistically}
\bibfield{author}{\bibinfo{person}{Amit Moryossef}, \bibinfo{person}{Zifan
  Jiang}, \bibinfo{person}{Mathias M{\"u}ller}, \bibinfo{person}{Sarah Ebling},
  {and} \bibinfo{person}{Yoav Goldberg}.} \bibinfo{year}{2023}\natexlab{}.
\newblock \bibinfo{title}{Linguistically {{Motivated Sign Language
  Segmentation}}}.
\newblock
\href{https://doi.org/10.48550/arXiv.2310.13960}{doi:\nolinkurl{10.48550/arXiv.2310.13960}}
\showeprint[arxiv]{2310.13960}~[cs]


\bibitem[{Naranjo-Zeled{\'o}n} et~al\mbox{.}(2020)]%
        {naranjo-zeledon-etal-2020-phonological}
\bibfield{author}{\bibinfo{person}{Luis {Naranjo-Zeled{\'o}n}},
  \bibinfo{person}{Mario {Chac{\'o}n-Rivas}}, \bibinfo{person}{Jes{\'u}s
  Peral}, {and} \bibinfo{person}{Antonio Ferr{\'a}ndez}.}
  \bibinfo{year}{2020}\natexlab{}.
\newblock \showarticletitle{Phonological {{Proximity}} in {{Costa Rican Sign
  Language}}}.
\newblock \bibinfo{journal}{\emph{Electronics}} \bibinfo{volume}{9},
  \bibinfo{number}{8} (\bibinfo{date}{Aug.} \bibinfo{year}{2020}),
  \bibinfo{pages}{1302}.
\newblock
\showISSN{2079-9292}
\href{https://doi.org/10.3390/electronics9081302}{doi:\nolinkurl{10.3390/electronics9081302}}


\bibitem[Nunnari et~al\mbox{.}(2024)]%
        {nunnari-etal-2024-dgsfabeln1}
\bibfield{author}{\bibinfo{person}{Fabrizio Nunnari},
  \bibinfo{person}{Eleftherios Avramidis}, \bibinfo{person}{Cristina
  {Espa{\~n}a-Bonet}}, \bibinfo{person}{Marco Gonz{\'a}lez},
  \bibinfo{person}{Anna Hennes}, {and} \bibinfo{person}{Patrick Gebhard}.}
  \bibinfo{year}{2024}\natexlab{}.
\newblock \showarticletitle{{{DGS-Fabeln-1}}: {{A Multi-Angle Parallel Corpus}}
  of {{Fairy Tales}} between {{German Sign Language}} and {{German Text}}}. In
  \bibinfo{booktitle}{\emph{Proceedings of the 2024 {{Joint International
  Conference}} on {{Computational Linguistics}}, {{Language Resources}} and
  {{Evaluation}} ({{LREC-COLING}} 2024)}},
  \bibfield{editor}{\bibinfo{person}{Nicoletta Calzolari},
  \bibinfo{person}{Min-Yen Kan}, \bibinfo{person}{Veronique Hoste},
  \bibinfo{person}{Alessandro Lenci}, \bibinfo{person}{Sakriani Sakti}, {and}
  \bibinfo{person}{Nianwen Xue}} (Eds.). \bibinfo{publisher}{{ELRA and ICCL}},
  \bibinfo{address}{Torino, Italia}, \bibinfo{pages}{4847--4857}.
\newblock
\urldef\tempurl%
\url{https://aclanthology.org/2024.lrec-main.434}
\showURL{%
\tempurl}


\bibitem[Renz et~al\mbox{.}(2021)]%
        {renz-etal-2021-sign}
\bibfield{author}{\bibinfo{person}{Katrin Renz}, \bibinfo{person}{Nicolaj~C.
  Stache}, \bibinfo{person}{Samuel Albanie}, {and}
  \bibinfo{person}{G{\textasciidieresis}ul Varol}.}
  \bibinfo{year}{2021}\natexlab{}.
\newblock \showarticletitle{Sign {{Language Segmentation}} with {{Temporal
  Convolutional Networks}}}.
\newblock \bibinfo{journal}{\emph{ICASSP, IEEE International Conference on
  Acoustics, Speech and Signal Processing - Proceedings}}
  \bibinfo{volume}{2021-June} (\bibinfo{year}{2021}),
  \bibinfo{pages}{2135--2139}.
\newblock
\showISSN{15206149}
\href{https://doi.org/10.1109/ICASSP39728.2021.9413817}{doi:\nolinkurl{10.1109/ICASSP39728.2021.9413817}}
\showeprint[arxiv]{2011.12986}


\bibitem[Ronchetti et~al\mbox{.}(2016)]%
        {lsa64-2016}
\bibfield{author}{\bibinfo{person}{Franco Ronchetti}, \bibinfo{person}{Facundo
  Quiroga}, \bibinfo{person}{C{\'e}sar~Armando Estrebou},
  \bibinfo{person}{Laura~Cristina Lanzarini}, {and} \bibinfo{person}{Alejandro
  Rosete}.} \bibinfo{year}{2016}\natexlab{}.
\newblock \showarticletitle{LSA64: an Argentinian sign language dataset}. In
  \bibinfo{booktitle}{\emph{XXII Congreso Argentino de Ciencias de la
  Computaci{\'o}n (CACIC 2016).}}
\newblock


\bibitem[Saiz and Garc{\'i}a(2023)]%
        {deespanola-2023-desafio}
\bibfield{author}{\bibinfo{person}{Mar{\'i}a Luz~Esteban Saiz} {and}
  \bibinfo{person}{Sa{\'u}l~Villameriel Garc{\'i}a}.}
  \bibinfo{year}{2023}\natexlab{}.
\newblock \bibinfo{booktitle}{\emph{{Informe Corpus de la Lengua de Signos
  Espa{\~n}ola (CORLSE): El desaf{\'i}o de documentar una lengua visual}}}.
\newblock \bibinfo{publisher}{Publicaciones Oficiales de la Administraci{\'o}n
  General del Estado}, \bibinfo{address}{Madrid}.
\newblock
\urldef\tempurl%
\url{https://www.corpuslse.es/corlse/consulta-y-acceso/informe_sociolinguistico_corlse_2023.pdf}
\showURL{%
\tempurl}


\bibitem[Tanzer(2025)]%
        {tanzer-2025-fleursasl}
\bibfield{author}{\bibinfo{person}{Garrett Tanzer}.}
  \bibinfo{year}{2025}\natexlab{}.
\newblock \showarticletitle{{{FLEURS-ASL}}: {{Including American Sign
  Language}} in {{Massively Multilingual Multitask Evaluation}}}. In
  \bibinfo{booktitle}{\emph{Proceedings of the 2025 {{Conference}} of the
  {{Nations}} of the {{Americas Chapter}} of the {{Association}} for
  {{Computational Linguistics}}: {{Human Language Technologies}} ({{Volume}} 1:
  {{Long Papers}})}}, \bibfield{editor}{\bibinfo{person}{Luis Chiruzzo},
  \bibinfo{person}{Alan Ritter}, {and} \bibinfo{person}{Lu~Wang}} (Eds.).
  \bibinfo{publisher}{Association for Computational Linguistics},
  \bibinfo{address}{Albuquerque, New Mexico}, \bibinfo{pages}{6167--6191}.
\newblock
\showISBNx{979-8-89176-189-6}
\href{https://doi.org/10.18653/v1/2025.naacl-long.314}{doi:\nolinkurl{10.18653/v1/2025.naacl-long.314}}


\bibitem[Tanzer and Zhang(2024)]%
        {tanzerzhang-2024-youtubesl25}
\bibfield{author}{\bibinfo{person}{Garrett Tanzer} {and} \bibinfo{person}{Biao
  Zhang}.} \bibinfo{year}{2024}\natexlab{}.
\newblock \bibinfo{title}{{{YouTube-SL-25}}: {{A Large-Scale}}, {{Open-Domain
  Multilingual Sign Language Parallel Corpus}}}.
\newblock
\href{https://doi.org/10.48550/arXiv.2407.11144}{doi:\nolinkurl{10.48550/arXiv.2407.11144}}
\showeprint[arxiv]{2407.11144}


\bibitem[{Trujillo-Romero} and {Garc{\'i}a-Bautista}(2023)]%
        {trujillo-romerogarcia-bautista-2023-mexican}
\bibfield{author}{\bibinfo{person}{Felipe {Trujillo-Romero}} {and}
  \bibinfo{person}{Gibran {Garc{\'i}a-Bautista}}.}
  \bibinfo{year}{2023}\natexlab{}.
\newblock \showarticletitle{Mexican {{Sign Language Corpus}}: {{Towards}} an
  {{Automatic Translator}}}.
\newblock \bibinfo{journal}{\emph{ACM Trans. Asian Low-Resour. Lang. Inf.
  Process.}} \bibinfo{volume}{22}, \bibinfo{number}{8} (\bibinfo{date}{Aug.}
  \bibinfo{year}{2023}), \bibinfo{pages}{212:1--212:24}.
\newblock
\showISSN{2375-4699}
\href{https://doi.org/10.1145/3591471}{doi:\nolinkurl{10.1145/3591471}}


\bibitem[Wilkinson et~al\mbox{.}(2016)]%
        {wilkinson-etal-2016-fair}
\bibfield{author}{\bibinfo{person}{Mark~D. Wilkinson}, \bibinfo{person}{Michel
  Dumontier}, \bibinfo{person}{IJsbrand~Jan Aalbersberg},
  \bibinfo{person}{Gabrielle Appleton}, \bibinfo{person}{Myles Axton},
  \bibinfo{person}{Arie Baak}, \bibinfo{person}{Niklas Blomberg},
  \bibinfo{person}{Jan-Willem Boiten}, \bibinfo{person}{Luiz~Bonino
  Da~Silva~Santos}, \bibinfo{person}{Philip~E. Bourne}, \bibinfo{person}{Jildau
  Bouwman}, \bibinfo{person}{Anthony~J. Brookes}, \bibinfo{person}{Tim Clark},
  \bibinfo{person}{Merc{\`e} Crosas}, \bibinfo{person}{Ingrid Dillo},
  \bibinfo{person}{Olivier Dumon}, \bibinfo{person}{Scott Edmunds},
  \bibinfo{person}{Chris~T. Evelo}, \bibinfo{person}{Richard Finkers},
  \bibinfo{person}{Alejandra {Gonzalez-Beltran}},
  \bibinfo{person}{Alasdair~J.G. Gray}, \bibinfo{person}{Paul Groth},
  \bibinfo{person}{Carole Goble}, \bibinfo{person}{Jeffrey~S. Grethe},
  \bibinfo{person}{Jaap Heringa}, \bibinfo{person}{Peter~A.C 'T~Hoen},
  \bibinfo{person}{Rob Hooft}, \bibinfo{person}{Tobias Kuhn},
  \bibinfo{person}{Ruben Kok}, \bibinfo{person}{Joost Kok},
  \bibinfo{person}{Scott~J. Lusher}, \bibinfo{person}{Maryann~E. Martone},
  \bibinfo{person}{Albert Mons}, \bibinfo{person}{Abel~L. Packer},
  \bibinfo{person}{Bengt Persson}, \bibinfo{person}{Philippe {Rocca-Serra}},
  \bibinfo{person}{Marco Roos}, \bibinfo{person}{Rene Van~Schaik},
  \bibinfo{person}{Susanna-Assunta Sansone}, \bibinfo{person}{Erik Schultes},
  \bibinfo{person}{Thierry Sengstag}, \bibinfo{person}{Ted Slater},
  \bibinfo{person}{George Strawn}, \bibinfo{person}{Morris~A. Swertz},
  \bibinfo{person}{Mark Thompson}, \bibinfo{person}{Johan Van Der~Lei},
  \bibinfo{person}{Erik Van~Mulligen}, \bibinfo{person}{Jan Velterop},
  \bibinfo{person}{Andra Waagmeester}, \bibinfo{person}{Peter Wittenburg},
  \bibinfo{person}{Katherine Wolstencroft}, \bibinfo{person}{Jun Zhao}, {and}
  \bibinfo{person}{Barend Mons}.} \bibinfo{year}{2016}\natexlab{}.
\newblock \showarticletitle{The {{FAIR Guiding Principles}} for Scientific Data
  Management and Stewardship}.
\newblock \bibinfo{journal}{\emph{Scientific Data}} \bibinfo{volume}{3},
  \bibinfo{number}{1} (\bibinfo{date}{March} \bibinfo{year}{2016}),
  \bibinfo{pages}{160018}.
\newblock
\showISSN{2052-4463}
\href{https://doi.org/10.1038/sdata.2016.18}{doi:\nolinkurl{10.1038/sdata.2016.18}}


\end{thebibliography}

\appendix

\end{document}